\pdfoutput=1

\documentclass[11pt]{article}

\usepackage{acl}

\usepackage{times}
\usepackage{latexsym}

\usepackage[T1]{fontenc}

\usepackage[utf8]{inputenc}

\usepackage{microtype}

\usepackage{inconsolata}

\usepackage{graphicx}

\usepackage{natbib}  
\usepackage{caption} 
\usepackage{algorithm}
\usepackage{algorithmic}
\usepackage{amsmath}
\usepackage{amssymb}
\usepackage{pifont}

\usepackage{colortbl}
\usepackage{multirow}
\usepackage{multicol}
\usepackage{float}

\usepackage{booktabs}
\usepackage{tcolorbox}
\tcbuselibrary{listingsutf8}

\newcommand{\cmark}{\ding{51}} 
\newcommand{\xmark}{\ding{55}} 

\definecolor{colorFst}{HTML}{bde6cd}       
\definecolor{colorSnd}{HTML}{e4eebc}       
\definecolor{colorTrd}{HTML}{fff8c5}       

\newcommand{\fs}{\cellcolor{colorFst}}   

\title{MaCP: Minimal yet Mighty Adaptation via Hierarchical Cosine Projection}

\author{
 \textbf{Yixian Shen}, 
    \textbf{Qi Bi}, 
    \textbf{Jia-Hong Huang}, 
    \textbf{Hongyi Zhu}, 
    \textbf{Andy D. Pimentel}, 
    \textbf{Anuj Pathania} \\
    University of Amsterdam, Amsterdam, the Netherlands \\
\\
 \texttt{\{y.shen, q.bi, j.huang, h.zhu, a.d.pimentel, a.pathania\}@uva.nl}
}

\begin{document}
\maketitle
\begin{abstract}
We present a new adaptation method MaCP, Minimal yet Mighty adaptive Cosine Projection, that achieves exceptional performance while requiring minimal parameters and memory for fine-tuning large foundation models.
Its general idea is to exploit the superior energy compaction and decorrelation properties of cosine projection to improve both model efficiency and accuracy.
Specifically, it projects the weight change from the low-rank adaptation into the discrete cosine space.
Then, the weight change is partitioned over different levels of the discrete cosine spectrum, and each partition's most critical frequency components are selected.
Extensive experiments demonstrate the effectiveness of MaCP across a wide range of single-modality tasks, including natural language understanding, natural language generation, text summarization, as well as multi-modality tasks such as image classification and video understanding. 
MaCP consistently delivers superior accuracy, significantly reduced computational complexity, and lower memory requirements compared to existing alternatives.
\end{abstract}

\section{Introduction}

Large language models (LLMs) have demonstrated exceptional performance across a wide range of natural language processing (NLP) tasks~\cite{huang2024novel,huang2024optimizing,huang2025image2text2image,wang2024performance,shi2025deep,shen2024comparative,shen2024altgen,du2025zero,tao2024robustness}, owing to their strong generalization ability achieved through training on large-scale corpora~\cite{touvron2023llama,liu2023gpt}. 
However, their zero-shot performance on many downstream tasks often remains suboptimal, failing to meet task-specific requirements. 
A common approach to address this limitation is full fine-tuning, where the model is adapted to a specific downstream task. However, the immense parameter size of LLMs makes this process computationally expensive and resource-intensive. 
For example, fine-tuning the state-of-the-art LLaMA 3.1~\cite{vavekanandllama} model with 70 billion parameters requires nearly 500GB of GPU memory, making it impractical for many real-world deployments~\cite{aghapour2024piqi,guo2024easter,shen2023thermal}.

Parameter-efficient fine-tuning (PEFT) methods effectively address these challenges. Key PEFT approaches include low-rank adaptation (LoRA)~\cite{hu2022lora}, adapter tuning~\cite{he2021towards}, and prompt tuning~\cite{lester2021power}. LoRA and its variants~\cite{liu2024dora,huang2025gradient} reduce trainable parameters by decomposing weight updates into two low-rank matrices, achieving SOTA accuracy. 
However, this efficiency in parameter count does not directly translate to reduced memory or compute cost. 
LoRA, for example, expands the effective embedding dimensionality, increasing FLOPs and requiring storage of high-dimensional activations and optimizer states during backpropagation~\cite{azizi2024lamda}. 
As summarized in Table~\ref{tab:comparison_methods}, existing PEFT techniques still suffer from partial inefficiencies, while methods like VeRA and LaMDA mitigate some memory overhead by optimizing specific components (e.g., gradients or optimizer states), 
they continue to incur significant costs in activation storage, overall memory footprint, and computational intensity.

\begin{table}[t!]
\centering
\scriptsize
\caption{Comparison of methods across memory optimization and computation intensity GFLOPs.}
\begin{tabular}{lcccc}
\toprule
\textbf{Method} & \multicolumn{3}{c}{\textbf{Memory}} & \textbf{GFLOPs} \\
\cmidrule(lr){2-4} 
& \textbf{Optimizer} & \textbf{Gradient} & \textbf{Activation} &  \\
\midrule
Full FT          & \xmark & \xmark & \xmark & Highest \\
LoRA             & \cmark & \cmark & \xmark & High \\
AdaLoRA          & \cmark & \cmark & \xmark & High \\
DoRA         & \cmark & \cmark & \cmark & High\\
VeRA             & \cmark & \cmark\cmark & \cmark &Low \\
AFLoRA           & \cmark & \cmark\cmark & \cmark & Medium \\
LaMDA & \cmark & \cmark\cmark & \cmark\cmark & Low \\
FourierFT             & \cmark & \cmark\cmark & \cmark\cmark & Low \\
\textbf{MaCP (Ours)} & \cmark\cmark & \cmark\cmark & \cmark\cmark & \textbf{Minimal} \\
\bottomrule
\end{tabular}
\label{tab:comparison_methods}
\end{table}

Recent research has begun exploring the frequency domain to compress and optimize model weights while preserving expressive capability. For instance, Gao et al.\cite{gao2024parameter} demonstrated that applying the Discrete Fourier Transform (DFT) to large language model (LLM) weight matrices can significantly reduce trainable parameters by focusing on the most informative spectral components, highlighting the promise of frequency-domain techniques for efficient fine-tuning. However, DFT is inherently suited for periodic signals\cite{nagai2020completing}, while the long-range dependencies in language context are often diverse and non-periodic~\cite{sun2021long,wen2022transformers}. Moreover, DFT operates in the complex domain, introducing computational overhead and potential numerical instability~\cite{press2007numerical}, particularly for large-scale models~\cite{gao2024parameter}. These numerical inaccuracies can degrade performance, making the energy compaction and stable adaptation of LLMs in the frequency domain less effective than intended. 

In this paper, we propose a novel fine-tuning approach that leverages the Discrete Cosine Transform (DCT) to address the existing bottlenecks. DCT is particularly well-suited for non-periodic signals and exhibits superior energy compaction and decorrelation properties, enabling a more efficient representation of critical information with fewer coefficients. To further enhance this efficiency, we partition the frequency domain using the discrete cosine coefficient matrix to identify and select the most informative components within the context. Specifically, MaCP simplifies the computational process compared to DFT-based approaches while preserving the expressive capacity of LLMs.

\begin{figure}
    \centering
    \includegraphics[width=0.9\linewidth]{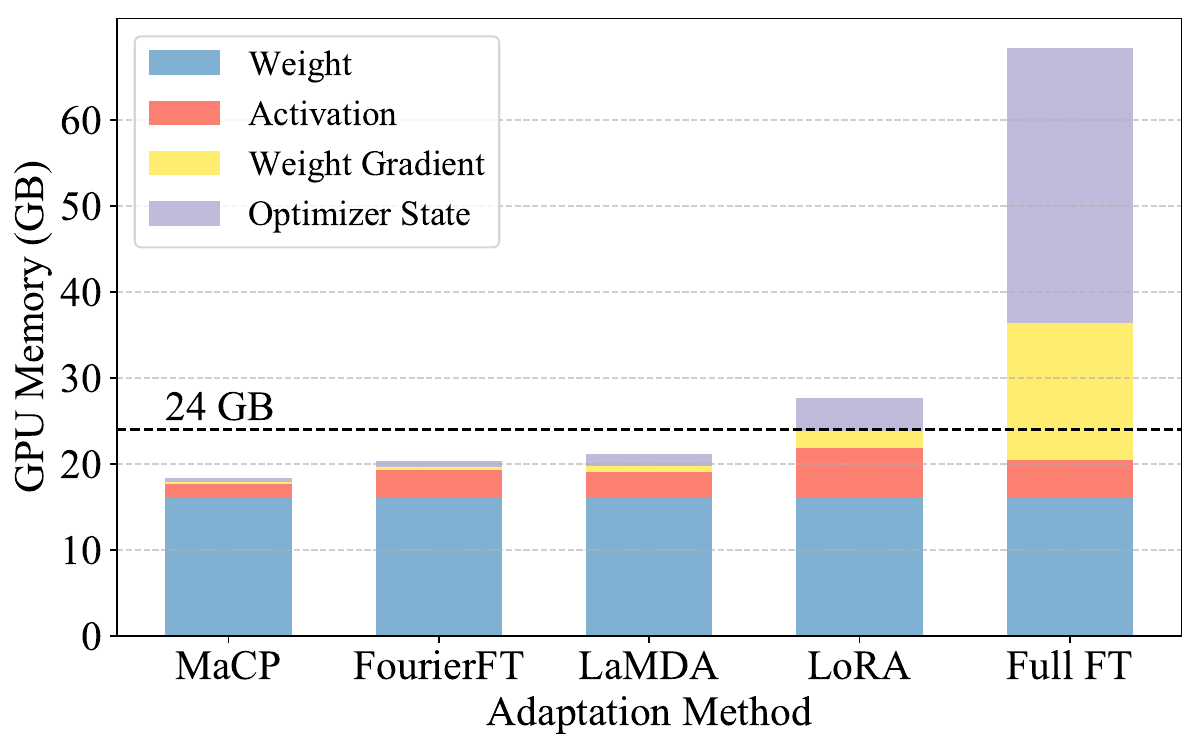}
    \caption{GPU memory usage of LLaMA3.1-8B on different fine-tuning methods}
    \label{fig:GPU}
\end{figure}

Building on these insights, by systematically selecting the most informative frequency components, compressing the weight updates, and then applying the inverse DCT (iDCT) to revert to the time domain, the proposed MaCP not only preserves model expressiveness but also drastically reduces memory and compute requirements.
As illustrated in Fig.\ref{fig:GPU}, MaCP minimizes the footprint of activations, gradients, and optimizer states, achieving significantly lower GPU memory usage than existing methods. By eliminating redundancy and reducing overhead across the training pipeline, MaCP offers a scalable and resource-efficient solution for fine-tuning large language models.

Our contribution can be summarized as follows.
\begin{itemize}
    \item We propose a novel Hierarchical Cosine Projection (MaCP) method for large language models, enabling minimal resource adaptation while preserving performance.
    \item We introduce an innovative frequency partitioning and selection strategy that exploits the Discrete Cosine Transform's (DCT) energy compaction property to enhance computational efficiency.
    \item We conduct extensive experiments on diverse benchmarks, demonstrating that MaCP outperforms existing PEFT methods in both performance and resource efficiency.
\end{itemize}

\section{Related Work}

\noindent \textbf{Low-Rank Adaptation} (LoRA)~\cite{hu2022lora} reduces trainable parameters by learning low-rank matrices, bypassing full-weight updates and minimizing memory usage for gradients and optimizers. Unlike adapter-based methods~\cite{he2021towards, pfeiffer2020adapterfusion, lin2020exploring, liao2023make, liao2023parameter,shen2025ssh,shen2024parameter}, LoRA incurs no inference overhead as low-rank updates merge with frozen weights. However, scaling LoRA to larger models and complex tasks remains challenging. Recent methods, including AdaLoRA~\cite{zhang2303adaptive}, VeRA~\cite{kopiczko2023vera}, QLoRA~\cite{dettmers2024qlora}, DoRA~\cite{liu2024dora}, and LaMDA~\cite{azizi2024lamda}, improve parameter allocation and decomposition but still face scalability issues on large models.

\noindent \textbf{Frequency-based Spectrum Learning} has been explored to reduce trainable parameters while maintaining model capacity. Prior works~\cite{xu2020learning,tang2022rethinking,yang2016exact} demonstrated the effectiveness of compact and sparse spectral representations. Gao et al.~\cite{gao2024parameter} applied the Fourier Transform to fine-tune a subset of spectral coefficients, showcasing the potential of sparse spectrum adaptation in large foundation models. However, DFT adds complex operations, and the asymmetry between the DFT and its inverse increases computational overhead.

\noindent \textbf{Discrete Cosine Transform}
has also been effectively applied in various deep learning contexts, demonstrating its potential for model compression and efficiency improvements. For instance, Xu et al. analyzed the spectral bias from a frequency perspective and proposed a learning-based frequency selection method to identify and remove trivial frequency components without loss of accuracy \cite{xu2020learning}. \citet{ehrlich2019deep}. leveraged DCT within the context of JPEG-compressed images, redefining convolution and batch normalization with a tunable approximation for ReLU operations \cite{ehrlich2019deep}. Furthermore, Zhu demonstrated that DCT could effectively recover data with very few parameters, highlighting its potential for model compression and efficiency \cite{zhu2024novel, cheng2024new}.

\section{Methodology}

\begin{algorithm}[tb]
\footnotesize
\caption{\textbf{MaCP} Algorithm}
\textbf{Input}: Input tensor $x$, number of parameters $n$, scaling factor $\alpha$, input dimension $d_1$, output dimension $d_2$, energy ratio $\delta$, pre-trained layer weights $W$\\
\textbf{Output}: Transformed tensor $h$
\begin{algorithmic}[1]
\STATE \textbf{Initialization:}
\STATE $W_{F} = \text{DCT}(base\_layer.weight)$  \textcolor{blue}{//DCT for  weights, Eq(\ref{eq1})}
\STATE \textcolor{blue}{// Partition frequencies into low, mid, and high components.}
\STATE $\mathcal{M}_{\text{low}}$, $\mathcal{M}_{\text{mid}}$, $\mathcal{M}_{\text{high}}$ $\leftarrow$ DistancePartition($W_{F}$) \textcolor{blue}{Eq(\ref{eq5},\ref{eq6},\ref{eq7})}
\STATE  \textcolor{blue}{// Select top-($n_\mathcal{M} \times \delta $ ) frequencies by energy}
\STATE $n_\mathcal{M} \times \delta $ $\leftarrow$ RankTopEnergyFreq($\mathcal{M}$)
\STATE \textcolor{blue}{// Randomly select the rest of frequencies}
\STATE $n_\mathcal{M} \times (1-\delta) $ $\leftarrow$ RandomSelectFreq($\mathcal{M}$)
\STATE \textcolor{blue}{// Stratified sampling from $\mathcal{M}_{\text{low}}$, $\mathcal{M}_{\text{mid}}$, $\mathcal{M}_{\text{high}}$, Eq(\ref{eq8})} 
\STATE $n \leftarrow n_{\mathcal{M}_\text{low}}+ n_{\mathcal{M}_\text{mid}} + n_{\mathcal{M}_\text{high}}$
\STATE \textcolor{blue}{// Initialize spectral coefficients}
\STATE $\Delta W_F$ $\leftarrow$ KaimingInitial()

\STATE \textbf{Forward Pass:}
\STATE \textcolor{blue}{// Set $n$ selected freq. trainable and froze ($d_1 \times d_2 - n$) freq.}
\STATE Set $n$ frequencies $\leftarrow$ requireGrad(True)
\STATE \textcolor{blue}{// Compute $\Delta W_T$ using inverse DCT based on Eq(\ref{eq9},\ref{eq10})}
\STATE $\Delta W_T \leftarrow \text{iDCT}(\Delta W_F) \times \alpha$
\STATE \textcolor{blue}{// Merge $\Delta W$ with base layer output} 
\STATE $h \leftarrow W + \Delta W_T$
\STATE \textbf{return} $h$
\end{algorithmic}
\label{alg:MaCP}
\end{algorithm}

\begin{figure}[!t]
    \centering
    \includegraphics[width=\linewidth]{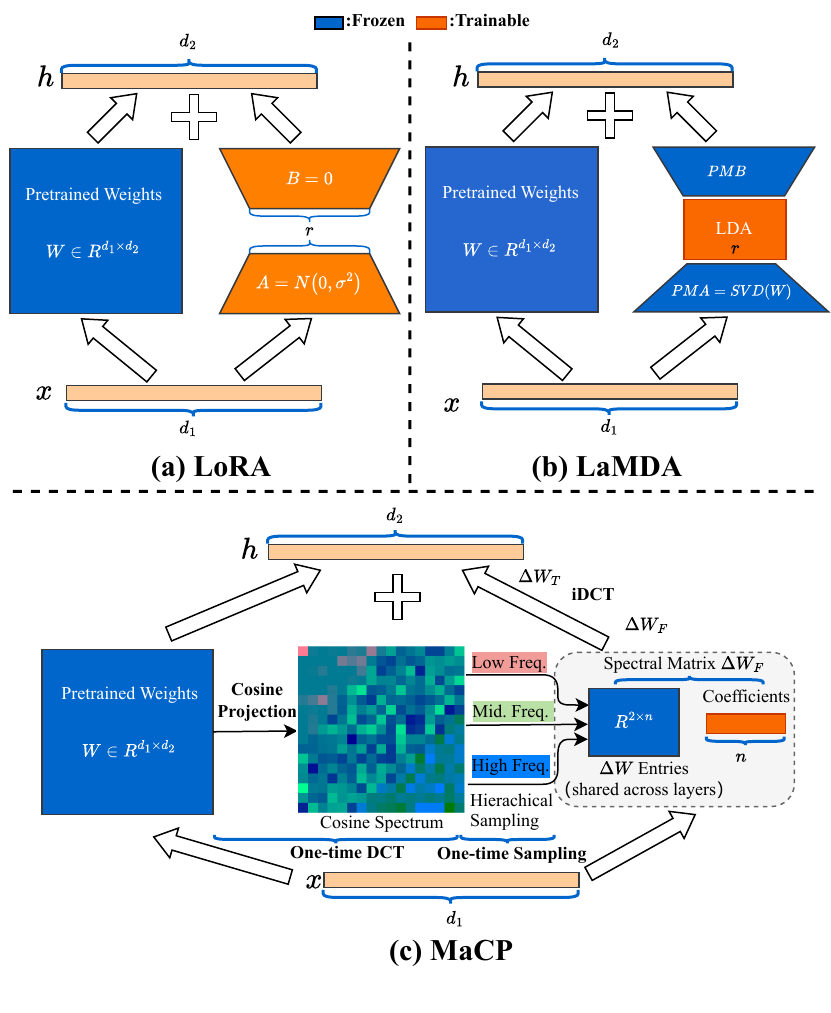}
    \caption{\small  (a) LoRA~\cite{hu2022lora} reduces trainable parameters by decomposing weight updates into low-rank matrices. (b) LaMDA~\cite{azizi2024lamda} further reduces trainable parameters through low-dimensional adaptation (LDA). (c) MaCP (ours) applies the Discrete Cosine Transform (DCT) to pre-trained weights, partitioning the frequency domain into low, middle, and high frequencies. A hierarchical sampling strategy then selects the most critical spectral coefficients for fine-tuning, reducing trainable parameters to $n$ per layer, which is significantly smaller compared to LoRA and LaMDA.
}
    \label{fig:MaCP}
\end{figure}

Fig.~\ref{fig:MaCP}~gives an overview of the proposed Selective Discrete Cosine Transform fine-tuning (MaCP) method. 
Different from the vanilla LoRA (low-rank adaptation) paradigm, MaCP operates in the frequency domain by learning a set of spectral coefficients on the cosine basis instead of directly modifying the spatial weights. 
The proposed MaCP identifies and retains the most critical frequency components by applying the DCT to the pre-trained weights. 
Then, it selects some spectral coefficients to learn trainable parameters, which form the spectral matrix. 
Finally, the iDCT is applied to the modified spectral matrix to transform it back to the spatial domain so as to update the weight changes, as detailed in Algorithm\ref{alg:MaCP}.


\subsection{Cosine Projection Using DCT}
The first step is to transform the weight matrices of the LLM into the frequency domain using the Discrete Cosine Transform (DCT). 
DCT is particularly advantageous due to its strong energy compaction properties, where the majority of the signal's energy is concentrated in a small number of low-frequency components. 
Specifically, given a weight matrix \( W[i, j] \) of size \( M \times N \), the 2D DCT is mathematically defined as:

\begin{small}
\begin{equation}
\label{eq1}
\begin{split}
&  W_F[u, v] = \ \alpha(u) \alpha(v) \sum_{i=0}^{M-1} \sum_{j=0}^{N-1} W[i, j] \\
& \times \cos\left[\frac{\pi}{M} \left(i + \frac{1}{2}\right) u\right] \cos\left[\frac{\pi}{N} \left(j + \frac{1}{2}\right) v\right],
\end{split}
\end{equation}
\end{small}

where \( u = 0, 1, \dots, M-1 \) and \( v = 0, 1, \dots, N-1 \) represent the frequency indices. The scaling factors \( \alpha(u) \) and \( \alpha(v) \) are defined as:
\begin{small}
\begin{equation}
\label{eq2}
\alpha(x) = 
\begin{cases}
\sqrt{\frac{1}{M}}, & \text{if } x = 0 \\
\sqrt{\frac{2}{M}}, & \text{if } x = 1, 2, \dots, M-1
\end{cases}
\end{equation}
\end{small}

This transformation converts the spatial (or time-domain) information of the weight matrix into a frequency-domain representation, where each element \( W_F[u, v] \) corresponds to a specific frequency component of the original matrix. Low-frequency components (where \( u \) and \( v \) are small) typically contain the most significant information, making them prime candidates for focused fine-tuning and significantly reducing the number of parameters that need to be fine-tuned.

\subsection{Hierarchical Cosine Spectrum}

The second step systematically partitions the frequency spectrum into three groups of distinct regions, namely, low, mid, and high frequencies. Each of these regions captures different aspects of the model's behavior, with low frequencies generally representing more global structures and high frequencies capturing finer.

 \textbf{Distance-Based Frequency Partitioning.}
We define the frequency domain as a 2D grid where each point \((u, v)\) corresponds to a specific frequency component in the Discrete Cosine Transform (DCT) of the weight matrix. The distance of each point from the origin (which represents the lowest frequency) is given by:

\begin{small}
\begin{equation}
\label{eq3}
d(u, v) = \sqrt{u^2 + v^2}
\end{equation} 
\end{small}
where \(u\) and \(v\) are the frequency indices corresponding to the horizontal and vertical directions in the DCT matrix, respectively.
The maximum distance from the origin is:

\begin{small}
\begin{equation}
\label{eq4}
d_{\text{max}} = \sqrt{\left(\frac{M}{2}\right)^2 + \left(\frac{N}{2}\right)^2}
\end{equation}   
\end{small}
where \(M\) and \(N\) are the dimensions of the DCT matrix. The frequency domain is partitioned into three distinct regions: low, medium, and high frequencies. These regions are defined based on their distance from the origin:

\textbf{Low-Frequency Components} (\(\mathcal{M}_{\text{low}}\)): Low frequencies are located close to the origin and are defined as:

\begin{small}
\begin{equation}
\label{eq5}
\mathcal{M}_{\text{low}} = \{(u, v) : d(u, v) \leq \frac{d_{\text{max}}}{3}\}
\end{equation}    
\end{small}

These components capture broad, global patterns and contain the majority of the energy due to the DCT's energy compaction property.

\textbf{Medium-Frequency Components} (\(\mathcal{M}_{\text{mid}}\)): Medium frequencies lie at an intermediate distance from the origin and are defined as:

\begin{small}
\begin{equation}
\label{eq6}
\mathcal{M}_{\text{mid}} = \{(u, v) : \frac{d_{\text{max}}}{3} < d(u, v) \leq \frac{2 \cdot d_{\text{max}}}{3}\}
\end{equation}
\end{small}
These components capture finer details and are essential for representing medium-scale structures.

\textbf{High-Frequency Components} (\(\mathcal{M}_{\text{high}}\)): High frequencies are located farthest from the origin and are defined as:

\begin{small}
\begin{equation}
\label{eq7}
\mathcal{M}_{\text{high}} = \{(u, v) : d(u, v) > \frac{2 \cdot d_{\text{max}}}{3}\}
\end{equation}
\end{small}
These components capture detailed features, which may include noise, but they are important for tasks requiring fine resolution.

Within each partition ( $\mathcal{M}_{\text{low}}$, $\mathcal{M}_{\text{mid}}$, $\mathcal{M}_{\text{high}}$), we apply a hybrid selection strategy that combines energy-based selection with a diversity-enhancing mechanism, where the top \( n_{\mathcal{M}_k} \times \delta \) coefficients are first selected based on energy (with \(\delta\) defaulting to 0.7 unless otherwise stated), followed by random selection for additional coefficients and stratified sampling to balance high-energy components and diverse frequencies across all partitions, in Algorithm~\ref{alg:MaCP} in Appendix. For more details on \(\delta\), see the ablation study in section~\ref{sec:exp}. The set of selected frequencies across all partitions is given by:

\begin{small}
\begin{equation}
\label{eq8}
n = \bigcup_{k=1}^{p} n_{\mathcal{M}_k},
\end{equation}    
\end{small}

where \( n_{\mathcal{M}_k} \) is the set of selected frequencies from partition \( \mathcal{M}_k \) and \( n_{\mathcal{M}_k} = n_{\text{energy},\mathcal{M}_k} + n_{\text{random},\mathcal{M}_k} \).

\subsection{Cosine Spectrum Learning via iDCT}

Once the most informative frequency components have been selected, the final step is to update the weights in the frequency domain and then transform them back to the space domain using the inverse Discrete Cosine Transform (iDCT). The updated weight matrix \( \Delta W \) is obtained via:

\begin{small}
\begin{equation}
\label{eq9}
\Delta W_T = \text{iDCT}(\Delta W_F),
\end{equation}
\end{small}

where \( \Delta W_F \) represents the modified frequency-domain weights, and iDCT is mathematically defined as:

\begin{small}
\begin{equation}
\label{eq10}
\begin{split}
& W[i, j] = \ \sum_{u=0}^{M-1} \sum_{v=0}^{N-1} \alpha(u) \alpha(v) W_F[u, v] \\
& \times \cos\left[\frac{\pi}{M} \left(i + \frac{1}{2}\right) u\right] \cos\left[\frac{\pi}{N} \left(j + \frac{1}{2}\right) v\right].
\end{split}
\end{equation} 
\end{small}

Note that only the selected frequency components in \( \Delta W_F \) are updated. The transformation back to the spatial domain integrates this fine-tuned information, preparing the model for inference or further training.

\subsection{Memory Efficiency Analysis}

The activation memory footprint of MaCP is significantly reduced because only a small subset of spectral coefficients $n$ is updated while keeping the remaining parameters frozen. The activation memory for MaCP consists of two components: (1) memory for the forward pass, requiring $B \cdot S \cdot H$ to store activations for all input-output pairs, where $S$ is the sequence length, $H$ is the embedding dimension, and $B$ is the batch size; and (2)memory for backpropagation, limited to the selected spectral coefficients, $B \cdot n$. Thus, the total activation memory footprint for MaCP is:

\begin{small}
\begin{equation}
    M_{act}^{MaCP} = B \cdot S \cdot H + B \cdot n,
\end{equation}
\end{small}

compared to LoRA, which requires:

\begin{small}
\begin{equation}
    M_{act}^{LoRA} = B \cdot S \cdot H + B \cdot S \cdot H,
\end{equation}
\end{small}

where $n < S \cdot H$, ensuring MaCP consumes less memory. For example, in a layer with $S = 2048$, $H = 4096$, $B = 1$, $n = 1000$, and $r = 32$ in LLaMA3.1-8B, reduces the activation memory footprint by $50.01$\% compared to LoRA, which underscore MaCP's efficiency, particularly in memory-constrained environments.

\begin{table*}[!ht]
\centering
\resizebox{0.77\textwidth}{!}{%
\begin{tabular}{cl|r|r|ccccccccc}
\toprule
& \textbf{Model} & \textbf{\# Trainable} & \textbf{MaCP's Para.}&\textbf{SST-2}$\uparrow$ & \textbf{MRPC}$\uparrow$ & \textbf{CoLA}$\uparrow$ & \textbf{QNLI}$\uparrow$ & \textbf{RTE}$\uparrow$ & \textbf{STS-B}$\uparrow$ & \multirow{2}{*}{\textbf{Avg.}} \\
& \textbf{\& Method} & \textbf{Parameters} &\textbf{Savings}& \textbf{(Acc.)} & \textbf{(Acc.)} & \textbf{(MCC)} & \textbf{(Acc.)} & \textbf{(Acc.)} & \textbf{(PCC)} \\
\midrule
\multirow{9}{*}{\rotatebox{90}{\textbf{BASE}}} 
& FF & 125M & \fs\textbf{99.99\%} & 94.8 & 90.2 & 63.6 & 92.8 & 78.7 & 91.2 & 85.22 \\ 
& BitFit & 0.1M &\fs\textbf{83.00\%} &  93.7 & \fs\textbf{92.7} & 62.0 & 91.8 & \fs\textbf{81.5} & 90.8 & \fs\textbf{85.42} \\ 
& Adpt\textsuperscript{D} & 0.9M & \fs\textbf{98.11\%}   & 94.7\scriptsize{$\pm$0.3} & 88.4\scriptsize{$\pm$0.1} & 62.6\scriptsize{$\pm$0.9} & 93.0\scriptsize{$\pm$0.2} & 75.9\scriptsize{$\pm$2.2} & 90.3\scriptsize{$\pm$0.1} & 84.15 \\
& LoRA & 0.3M & \fs\textbf{94.33\%}   & \fs\textbf{95.1}\scriptsize{$\pm$0.2} & 89.7\scriptsize{$\pm$0.7} & 63.4\scriptsize{$\pm$1.2} & \fs\textbf{93.3}\scriptsize{$\pm$0.3} & 78.4\scriptsize{$\pm$0.8} & \fs\textbf{91.5}\scriptsize{$\pm$0.2} & 85.23 \\ 
& AdaLoRA & 0.3M & \fs\textbf{94.33\%} & 94.5\scriptsize{$\pm$0.2} & 88.7\scriptsize{$\pm$0.5} & 62.0\scriptsize{$\pm$0.6} & 93.1\scriptsize{$\pm$0.2} & 81.0\scriptsize{$\pm$0.6} & 90.5\scriptsize{$\pm$0.2} & 84.97 \\
& DoRA & 0.3M &  \fs\textbf{94.33\%} &94.9\scriptsize{$\pm$0.5} & 89.9\scriptsize{$\pm$0.5} & 63.7\scriptsize{$\pm$1.1} & \fs\textbf{93.3}\scriptsize{$\pm$0.2} & 78.9\scriptsize{$\pm$0.6} & \fs\textbf{91.5}\scriptsize{$\pm$0.2} & 85.37 \\ 
& VeRA & 0.043M &  \fs\textbf{60.47\%} &94.6\scriptsize{$\pm$0.1} & 89.5\scriptsize{$\pm$0.5} & \textbf{65.6}\scriptsize{$\pm$0.9} & 91.8\scriptsize{$\pm$0.3} & 78.7\scriptsize{$\pm$0.7} & 90.7\scriptsize{$\pm$0.2} & 85.15 \\
& LaMDA & 0.089M &  \fs\textbf{80.90\%} &94.5\scriptsize{$\pm$0.2} & 89.3\scriptsize{$\pm$0.3} & \fs\textbf{65.6}\scriptsize{$\pm$0.5} & 91.2\scriptsize{$\pm$0.6} & 78.9\scriptsize{$\pm$0.3} & 90.5\scriptsize{$\pm$0.3} & 85.01 \\
& FourierFT & 0.024M & \fs\textbf{29.17\%} &94.2\scriptsize{$\pm$0.3} & 90.0\scriptsize{$\pm$0.8} & 63.8\scriptsize{$\pm$1.6} & 92.2\scriptsize{$\pm$0.1} & 79.1\scriptsize{$\pm$0.5} & 90.8\scriptsize{$\pm$0.2} & 85.02 \\ 
& \textbf{MaCP} & \fs\textbf{0.017M} &-& 94.2\scriptsize{$\pm$0.2} & 89.7\scriptsize{$\pm$0.7} & 64.6\scriptsize{$\pm$1.4} & 92.4\scriptsize{$\pm$0.1} & 80.7\scriptsize{$\pm$0.9} & 90.9\scriptsize{$\pm$0.3} & \fs\textbf{85.42} \\ 
\midrule
\multirow{8}{*}{\rotatebox{90}{\textbf{LARGE}}} 
& FF & 356M &\fs\textbf{99.99\%} &  96.3 & 90.9 & 68.0 & 94.7 & 86.6 & 92.4 & 88.11 \\ 
& Adpt\textsuperscript{P} & 3M & \fs\textbf{98.87\%}   & 96.1\scriptsize{$\pm$0.3} & 90.2\scriptsize{$\pm$0.7} & \fs\textbf{68.3}\scriptsize{$\pm$1.0} & 94.7\scriptsize{$\pm$0.2} & 83.8\scriptsize{$\pm$2.9} & 92.1\scriptsize{$\pm$0.7} & 87.55 \\ 
& Adpt\textsuperscript{P} & 0.8M  & \fs\textbf{	95.75\%}& \fs\textbf{96.6}\scriptsize{$\pm$0.2} & 89.7\scriptsize{$\pm$1.2} & 67.8\scriptsize{$\pm$2.5} & 94.7\scriptsize{$\pm$0.3} & 80.1\scriptsize{$\pm$2.9} & 91.9\scriptsize{$\pm$0.4} & 86.82 \\ 
& Adpt\textsuperscript{H} & 6M  & \fs\textbf{99.43\%}& 96.2\scriptsize{$\pm$0.3} & 88.7\scriptsize{$\pm$2.9} & 66.5\scriptsize{$\pm$4.4} & 94.7\scriptsize{$\pm$0.2} & 83.4\scriptsize{$\pm$1.1} & 91.0\scriptsize{$\pm$1.7} & 86.75 \\
& Adpt\textsuperscript{H} & 0.8M  & \fs\textbf{95.75\%}& 96.3\scriptsize{$\pm$0.5} & 87.7\scriptsize{$\pm$1.7} & 66.3\scriptsize{$\pm$2.0} & 94.7\scriptsize{$\pm$0.2} & 72.9\scriptsize{$\pm$2.9} & 91.5\scriptsize{$\pm$0.5} & 84.90 \\ 
& LoRA & 0.8M  & \fs\textbf{95.75\%}& 96.2\scriptsize{$\pm$0.5} & 90.2\scriptsize{$\pm$1.0} & 68.2\scriptsize{$\pm$1.9} & \fs\textbf{94.8}\scriptsize{$\pm$0.3} & 85.2\scriptsize{$\pm$1.1} & 92.3\scriptsize{$\pm$0.5} & 87.82 \\
& DoRA & 0.9M  & \fs\textbf{96.22\%}& 96.4\scriptsize{$\pm$0.4} & \fs\textbf{91.0}\scriptsize{$\pm$1.1} & 67.2\scriptsize{$\pm$1.6} & \fs\textbf{94.8}\scriptsize{$\pm$0.3} & 85.4\scriptsize{$\pm$1.1} & 92.1\scriptsize{$\pm$0.3} & 87.82 \\
& VeRA & 0.061M &\fs\textbf{44.26\%}& 96.1\scriptsize{$\pm$0.1} & 90.9\scriptsize{$\pm$0.7} & 68.0\scriptsize{$\pm$1.8} & 94.4\scriptsize{$\pm$0.2} & 85.9\scriptsize{$\pm$0.7} & 91.7\scriptsize{$\pm$0.8} & 87.83 \\
& LaMDA & 0.12M &\fs\textbf{71.67\%}& 96.0\scriptsize{$\pm$0.1} & 90.9\scriptsize{$\pm$0.2} & 67.1\scriptsize{$\pm$1.2} & 94.2\scriptsize{$\pm$0.5} & 85.3\scriptsize{$\pm$0.5} & 91.3\scriptsize{$\pm$0.4} & 87.47 \\
& FourierFT & 0.048M &\fs\textbf{29.17\%}& 96.0\scriptsize{$\pm$0.2} & 90.9\scriptsize{$\pm$0.3} & 67.1\scriptsize{$\pm$1.4} & 94.4\scriptsize{$\pm$0.4} & \fs\textbf{87.4}\scriptsize{$\pm$1.6} & 91.9\scriptsize{$\pm$0.4} & 87.95 \\ 
& \textbf{MaCP} & \fs\textbf{0.034M} & -&96.2\scriptsize{$\pm$0.3} & 90.9\scriptsize{$\pm$0.3} & 67.7\scriptsize{$\pm$1.4} & 94.5\scriptsize{$\pm$0.5} & \fs\textbf{87.4}\scriptsize{$\pm$1.1} & \fs\textbf{92.0}\scriptsize{$\pm$0.5} & \fs\textbf{88.12} \\ 
\bottomrule
\end{tabular}%
}
\caption{\small Performance of various fine-tuning methods on 6 GLUE tasks, using base and large models. Metrics include MCC for CoLA, PCC for STS-B, and accuracy for other tasks. Results are medians of 5 runs with different seeds; the best scores in each category are bolded. }
\label{tab:nlup}
\end{table*}

\section{Experiments}
\label{sec:exp}

\subsection{Baselines}
The proposed MaCP is compared with the following baselines: \textbf{Full Fine-Tuning (FF)} updates all parameters of the model. \textbf{Adapter Tuning}~\cite{houlsby2019parameter,lin2020exploring,ruckle2020adapterdrop,pfeiffer2020adapterfusion} introduces adapter layers between the self-attention and MLP modules. \textbf{LoRA}~\cite{hu2022lora} approximates weight updates using low-rank matrices to minimize trainable parameters, while \textbf{AdaLoRA}~\cite{zhang2303adaptive} dynamically allocates the parameter budget based on importance scores. \textbf{DoRA}~\cite{liu2024dora} decomposes pre-trained weights into magnitude and direction, applying LoRA for directional updates. \textbf{VeRA}~\cite{kopiczko2023vera} uses a single pair of low-rank matrices across all layers with small scaling vectors to reduce parameters while maintaining performance. \textbf{FourierFT}~\cite{gao2024parameter} fine-tunes a subset of spectral coefficients in the Fourier domain, and \textbf{LaMDA}~\cite{azizi2024lamda} employs low-dimensional adaptation to further reduce trainable parameters.

\subsection{Natural Language Understanding}

\noindent \textbf{Models, Datasets, and Implementation.} We evaluate MaCP on the GLUE benchmark~\cite{wang2019glue} using RoBERTa~\cite{liu2019roberta} in both Base and Large configurations, covering diverse NLU tasks for comprehensive assessment. \textbf{Implementation Details.} MaCP uses 700 out of 768\textsuperscript{2} spectral coefficients for RoBERTa Base and 1024\textsuperscript{2} for RoBERTa Large, selecting the most critical components uniformly across all layers. To ensure fair comparison, we adopt the same experimental settings as LoRA, with hyperparameters provided in the appendix.

\textbf{Results and Analysis}
The results are presented in Tab.~\ref{tab:nlup}, where we report the median performance across 5 random seed runs, with the best epoch selected for each run. The proposed MaCP consistently delivers superior or comparable performance to baseline methods while significantly reducing the number of trainable parameters. For instance, MaCP achieved an 80.7\% accuracy on RTE and a 92.0 Pearson correlation on STS-B, outperforming methods like LoRA and FourierFT with significantly fewer parameters. On SST-2, MaCP maintained a high accuracy of 96.2\%, matching or slightly outperforming other approaches. Additionally, it recorded a 94.5\% accuracy on QNLI, further highlighting its effectiveness. Across all tasks, MaCP shows robust generalization while requiring minimal training parameters.


\begin{table}[!ht]
\centering
\scalebox{0.6}{
\begin{tabular}{l|lr|crcccccc}
\toprule
 & \textbf{Method} & \textbf{\# Tr. Para.} & \textbf{BLEU} & \textbf{NIST} & \textbf{METE.} & \textbf{ROU-L} & \textbf{CIDEr} \\
\midrule
\multirow{9}{*}{\rotatebox{90}{\textbf{GPT-2 Medium}}} 
& FT\textsuperscript{1} & 354.92M & 68.2 & 8.62 & 46.2 & 71.0 & 2.47 \\
& Adpt\textsuperscript{L\textsuperscript{1}} & 0.37M & 66.3 & 8.41 & 45.0 & 69.8 & 2.40 \\
& Adpt\textsuperscript{L\textsuperscript{1}} & 11.09M & 68.9 & 8.71 & 46.1 & 71.3 & 2.47 \\
& Adpt\textsuperscript{H\textsuperscript{1}} & 11.09M & 67.3 & 8.50 & 46.0 & 70.7 & 2.44 \\
& LoRA & 0.35M & 68.9 & 8.76 & 46.6 & 71.5 & 2.51 \\
& DoRA & 0.36M & 69.2 & 8.79 & 46.9 & 71.7 & 2.52\\
& VeRA & 0.35M & \fs\textbf{70.1} & 8.81 & 46.6 & 71.5 & 2.50 \\
& FourierFT & 0.048M & 69.1 & \fs\textbf{8.82} & 47.0 & 71.8 & 2.51 \\
& \textbf{MaCP} & \fs\textbf{0.034M} & \fs\textbf{70.1} & \fs\textbf{8.82} & \fs\textbf{47.2} & \fs\textbf{71.9} & \fs\textbf{2.53} \\
\midrule
\multirow{8}{*}{\rotatebox{90}{\textbf{GPT-2 Large}}} 
& FT\textsuperscript{1} & 774.03M & 68.5 & 8.78 & 46.0 & 69.9 & 2.45 \\
& Adpt\textsuperscript{L\textsuperscript{1}} & 0.88M & 69.1 & 8.68 & 46.1 & 71.0 & 2.49 \\
& Adpt\textsuperscript{L\textsuperscript{1}} & 23.00M & 68.9 & 8.70 & 46.1 & 71.3 & 2.45 \\
& LoRA & 0.77M & 69.4 & 8.81 & 46.5 & \fs\textbf{71.9} & 2.50 \\
& DoRA & 0.79M & 69.8 & 8.83 & 46.9 & \fs\textbf{71.9} & 2.50 \\
& VeRA & 0.17M & \fs\textbf{70.3} & 8.85 & 46.6 & 71.6 & 2.54 \\
& FourierFT & 0.072M & 70.2 & 8.90 & 47.0 & 71.8 & 2.50 \\
& \textbf{MaCP} & \fs\textbf{0.051M} & \fs\textbf{70.3} & \fs\textbf{8.92} & \fs\textbf{47.1} & \fs\textbf{71.9} & \fs\textbf{2.55} \\
\bottomrule
\end{tabular}}
\caption{\small Performance comparison of different fine-tuning methods on GPT-2 Medium and GPT-2 Large models, evaluated using BLEU, NIST, METEOR, ROUGE-L, and CIDEr metrics. \textsuperscript{1} denotes results sourced from previous studies. The best score for each metric is highlighted in bold.}
\label{tab:e2e}
\end{table}

\subsection{Natural Language Generation}
\noindent \textbf{Models and Datasets.}
We evaluate MaCP on the E2E natural language generation (NLG) task~\cite{novikova2017e2e} by fine-tuning GPT-2 (Medium and Large)~\cite{radford2019language} models, which are decoder-only architectures with 24 and 36 transformer blocks.

\noindent \textbf{Implementation Details.} 
We fine-tune LoRA, DoRA, FourierFT, VeRA, and the proposed MaCP method on the GPT-2 Medium and Large models using a linear learning rate scheduler over 5 epochs. Results are averaged over 3 runs, and detailed hyperparameters are in the Appendix.

\noindent \textbf{Results and Analysis.} Tab.~\ref{tab:e2e} shows that MaCP consistently outperforms other methods across most metrics while requiring significantly fewer trainable parameters. Specifically, MaCP reduces trainable parameters by 9.59\% and 6.6\% compared to LoRA for GPT-2 Medium and Large models, respectively, and achieves a 30\% parameter reduction compared to FourierFT.

\begin{table}
\centering
\resizebox{0.43\textwidth}{!}{%
\begin{tabular}{l|l|c|crcc}
\toprule
\textbf{Model} & \textbf{Method} & \textbf{\# Tr. Para.} & \textbf{MT-Bench} & \textbf{Vicuna} \\
\midrule
\multirow{5}{*}{\textbf{LLaMA2-7B}} 
& LoRA & 159.9M & 5.19 & 6.85 \\
& DoRA & 163.7M & 5.20 & 6.86 \\
& VeRA & 1.6M & 5.18 & 6.82 \\
& FourierFT & 0.064M & 5.09 & 6.85 \\
& \textbf{MaCP} & \fs\textbf{0.045M} & \fs\textbf{5.22} & \fs\textbf{6.87}\\
\midrule
\multirow{5}{*}{\textbf{LLaMA2-13B}} 
& LoRA & 250.3M & 5.77 & 7.38\\
& DoRA & 264.5M & 5.79 & 7.47 \\
& VeRA & 2.4M & \fs\textbf{5.93} & 7.41 \\
& FourierFT & 0.08M & 5.82 & 7.49 \\
& \textbf{MaCP} & \fs\textbf{0.056M} & \fs\textbf{5.93} & \fs\textbf{7.55} \\
\midrule
\multirow{5}{*}{\textbf{LLaMA3.1-8B}} 
& LoRA & 183.3M & 5.65 & 7.11 \\
& DoRA & 186.9M & 5.66 & \fs\textbf{7.19} \\
& VeRA & 1.9M & 5.61 & 7.13 \\
& FourierFT & 0.064M & 5.67 & 7.15 \\
& \textbf{MaCP} & \fs\textbf{0.05M} & \fs\textbf{5.69} & \fs\textbf{7.19} \\
\bottomrule
\end{tabular}%
}
\caption{\small The average scores on MT-Bench and Vicuna assessed by GPT-4. The higher score is better.}
\label{tab:mtbench_vicuna}
\end{table}

\subsection{Text Summarization}
\noindent \textbf{Models and Datasets.}  
We assess MaCP's effectiveness on text summarization using the BART-Large model~\cite{lewis2019bart}, evaluated on two standard benchmarks: XSUM~\cite{narayan2018don} and CNN/DailyMail~\cite{hermann2015teaching}.

\noindent \textbf{Implementation Details.}  
We compare MaCP with SOTA baselines under consistent settings. 
For LaMDA, we use a rank of $r=32$, and for MaCP, we select $n=5000$ cosine spectrum coefficients. 
All models are trained with a learning rate of $2\times10^{-4}$; batch sizes are 32 for XSUM and 64 for CNN/DailyMail. 
Training runs for 25 epochs on XSUM and 15 on CNN/DailyMail.

\noindent \textbf{Results and Analysis.}  
Table~\ref{tab:nlg_bart} shows that MaCP achieves the best overall performance on both XSUM and CNN/DailyMail, despite using drastically fewer trainable parameters.
On XSUM, MaCP outperforms all baselines with ROUGE-1/2/L scores of 45.21/22.19/37.10, even surpassing full fine-tuning, which requires over 2400$\times$ more parameters. 
Similarly, on CNN/DailyMail, MaCP delivers top results with 45.09/21.97/42.29, outperforming all other PEFT baselines. These results demonstrate that MaCP preserves key semantic and structural information during fine-tuning, while operating under extreme parameter efficiency.

\noindent Notably, MaCP’s advantage over FourierFT underscores the benefit of using cosine over Fourier bases. DCT's real-valued, non-periodic decomposition aligns better with the structure of language data, enabling more effective compression and adaptation.
Compared to LaMDA and VeRA, which also aim for compactness, MaCP delivers significantly better ROUGE scores while using fewer parameters. 


\begin{table}[h!]
\scriptsize
\centering
\setlength{\tabcolsep}{2pt} 
\begin{tabular}{l|ccc}
\toprule
\textbf{Method} & \textbf{\#Param.} & \textbf{XSUM} & \textbf{CNN/DailyMail} \\
\midrule
Full fine-tuning       & 415M & 45.14/22.27/37.25 & 44.16/21.28/40.90 \\
LoRA                   & 8.6M & 43.95/20.72/35.68 & 45.03/21.84/42.15 \\
DoRA                 & 8.8M & 44.23/21.17/36.02 & 44.92/21.89/42.09 \\
VeRA        & 0.92M & 43.72/20.33/35.11 & 44.87/21.26/40.25 \\
LaMDA      & 0.85M & 43.92/20.68/35.21 & 44.12/21.16/40.45 \\
Fourier      & 0.33M & 44.39/20.91/36.05 & 44.42/21.07/40.53 \\
MaCP  & \fs \textbf{0.17M} & \fs\textbf{45.21/22.19/37.10} & \fs\textbf{45.09/21.97/42.29} \\
\bottomrule
\end{tabular}
\caption{\small Comparison of fine-tuning methods on XSUM and CNN/DailyMail benchmarks. Bold highlights the best results.}
\label{tab:nlg_bart}
\end{table}

\subsection{Instruction Tuning}

\noindent \textbf{Models, Datasets, and Implementation.}  
We fine-tune LLaMA2-7B, LLaMA2-13B, and LLaMA3.1-8 using MaCP and baseline methods on the Alpaca dataset~\cite{taori2023stanford}. For evaluation, responses are generated for pre-defined questions from MT-Bench~\cite{zheng2024judging} and Vicuna Eval datasets, and scored by GPT-4 on a 10-point scale. Following prior work~\cite{dettmers2024qlora,dettmers20228bit}, LoRA, DoRA, and VeRA are applied to all linear layers except the top one, while FourierFT follows the settings in~\cite{gao2024parameter}. For MaCP, we set \(n=700\). To train on a single GPU, we adopt the QLoRA quantization technique~\cite{dettmers2024qlora}. All methods are trained for one epoch, and the average scores of all responses are reported. Hyperparameter details are provided in the Appendix.

\noindent \textbf{Results and Analysis.}
Tab.~\ref{tab:mtbench_vicuna} show that the MaCP consistently outperforms or matches baseline methods across different LLaMA models. 
It achieves the highest scores on both MT-Bench and Vicuna for LLaMA2-7B and LLaMA2-13B, demonstrating its effectiveness in generating high-quality responses with significantly fewer trainable parameters. 
For LLaMA2-7B, MaCP requires only 0.045M parameters, reducing 99.7\% compared to LoRA, yet it delivers superior performance. 
Similarly, for LLaMA2-13B, MaCP matches the top score of 5.93 on MT-Bench and achieves the highest score of 7.55 on Vicuna, with only 0.056M parameters. 
For LLaMA3.1-8B, MaCP shows competitive results, outperforming other methods with a score of 5.69 on MT-Bench and matching the top score of 7.19 on Vicuna, with minimal trainable parameters.

\begin{table}
\centering
\resizebox{0.47\textwidth}{!}{%
\begin{tabular}{l|l|r|cccccc}
\toprule
\textbf{Model} & \textbf{Method} & \textbf{\# Train. Para.} & \textbf{CIFAR100} & \textbf{DTD} & \textbf{EuroSAT} & \textbf{OxfordPets} \\
\midrule
\multirow{7}{*}{\textbf{ViT-B}} 
& Head & - & 84.3 & 69.8 & 88.7 & 90.3 \\
& Full & 85.8M & \fs\textbf{92.4} & \fs\textbf{77.7} & \fs\textbf{99.1}& \fs\textbf{93.4} \\
& LoRA & 581K & 92.1 & 75.2 & 98.4 & 93.2 \\
& DoRA & 594K & 92.3 & 75.3 & 98.7 & 93.2 \\
& VeRA & 57.3K & 91.7 & 74.6 & 98.5 & \fs\textbf{93.4}\\
& FourierFT & 72K & 91.2 & 75.1 & 98.8 & 93.2 \\
& \textbf{MaCP} & \fs\textbf{50.4K} & 91.7 & 75.9 & \fs\textbf{99.1} & \fs\textbf{93.4} \\

\midrule
\multirow{7}{*}{\textbf{ViT-L}} 
& Head & - & 84.7 & 73.3 & 92.6 & 91.1 \\
& Full & 303.3M & 93.6 & 81.8 & \fs\textbf{99.1} & 94.4 \\
& LoRA & 1.57M &  94.9 & 81.8 & 98.63 & \fs\textbf{94.8} \\
& DoRA & 1.62M &  \fs\textbf{95.1} & 81.8 &   98.8 & \fs\textbf{94.8}\\
& VeRA & 130.5K & 94.2 & 81.6& 98.6 & 93.7 \\
& FourierFT & 144K & 93.7 & 81.2 & 98.7 & 94.5 \\
& \textbf{MaCP} & \fs\textbf{100.9K} & 94.5 & \fs\textbf{81.9} & 99.0& \fs\textbf{94.8} \\
\bottomrule
\end{tabular}%
}
\caption{\small Performance of various fine-tuning methods on ViT-B and ViT-L models across different datasets. The best results for each dataset are highlighted in bold.}
\label{tab:vit_results}
\end{table}

\subsection{Image Classification}

\noindent \textbf{Models, Datasets, and Implementation.}  
We evaluate our method on the Vision Transformer (ViT)~\cite{dosovitskiy2020image} in Base and Large variants using CIFAR-100~\cite{krause20133d}, DTD~\cite{cimpoi2014describing}, EuroSAT~\cite{helber2019eurosat}, and OxfordPets~\cite{parkhi2012cats}. LoRA, DoRA, VeRA, and six other baselines are applied to the query and value layers of ViT, while training only the classification head is denoted as "Head." We set \(r = 16\) for LoRA, \(n = 3000\) for FourierFT, and \(n = 2400\) for MaCP, with learning rates and weight decay tuned for all methods. Training is capped at 10 epochs, with hyperparameter details provided in the Appendix.

\noindent \textbf{Results and Analysis.}
Tab.~\ref{tab:vit_results} presents the performance on ViT-B and ViT-L across four image classification datasets. 
For the ViT-B model, MaCP demonstrates competitive performance with only 50.4K trainable parameters, while LoRA and DoRA require over 10 times more parameters. Specifically, MaCP matches the full fine-tuning performance on EuroSAT and OxfordPets, achieving 99.1\% and 93.4\% accuracy, respectively. 
For the ViT-L model, MaCP achieves near-optimal performance with only 100.9K parameters.
MaCP records the best result on DTD at 81.9\%, while matching the top accuracy on OxfordPets at 94.8\%.

\begin{table}[!ht]
\centering
\resizebox{0.47\textwidth}{!}{%
\begin{tabular}{l|l|r|cccccc}
\toprule
\textbf{Model} & \textbf{Method} & \textbf{\# Train. Para.} & \textbf{TVQA} & \textbf{How2QA} & \textbf{TVC} & \textbf{YC2C} \\
\midrule
\multirow{7}{*}{\textbf{VL-BART}} 
& Full & 228.9M & \fs\textbf{76.3} & 73.9 & 45.7& \fs\textbf{154} \\
& LoRA & 11.8M & 75.5 & 72.9 & 44.6 & 140.9 \\
& DoRA & 11.9M & \fs\textbf{76.3} & 74.1 & 45.8 & 145.4 \\
& VeRA & 1.3M & 75.9 & 73.8 & 44.7 & 142.6\\
& FourierFT & 1.5M & 76.2 & 73.1 & 45.5 & 147.3 \\
& \textbf{MaCP} & \fs\textbf{1.1M} & \fs\textbf{76.3} & \fs\textbf{74.3}& \fs\textbf{45.9} & \fs\textbf{154} \\

\bottomrule
\end{tabular}%
}

\caption{\small Multi-task evaluation results on TVQA, How2QA, TVC, and YC2C using the VL-BART backbone. MaCP achieves top or near-top scores across all tasks with significantly fewer trainable parameters, demonstrating its efficiency and effectiveness.}

\label{tab:bart_results}
\end{table}

\subsection{Video-Text Understanding}

\noindent \textbf{Models, Datasets, and Implementation.}  
We compare DoRA, LoRA, and full fine-tuning on VL-BART, which integrates CLIP-ResNet101~\cite{radford2021learning} as the vision encoder and BART-Base~\cite{lewis2019bart} as the language model. The evaluation covers four tasks: TVQA~\cite{lei2018tvqa} and How2QA~\cite{li2020hero} for video question answering, and TVC~\cite{lei2020tvr} and YC2C~\cite{zhou2018towards} for video captioning. Following~\cite{sung2022vl}, VL-BART is fine-tuned in a multi-task setup. We set \( r = 128 \) for LoRA, \( n = 6000 \) for FourierFT, and \( n = 4500 \) for MaCP, with tuned learning rates and weight decay, training for 7 epochs.


\noindent \textbf{Results and Analysis.}  
Table~\ref{tab:bart_results} presents the multi-task evaluation results for video-text tasks using the VL-BART backbone. MaCP achieves top or near-top performance across all tasks while using significantly fewer trainable parameters. For TVQA and YC2C, MaCP matches the highest scores of \textbf{76.3} and \textbf{154}, respectively, while reducing parameters by over 99\% compared to full fine-tuning. On How2QA, MaCP records the highest score of \textbf{74.3}, outperforming all baselines. For TVC, MaCP achieves a strong result of \textbf{45.9}, slightly surpassing DoRA and FourierFT, further demonstrating its efficiency and effectiveness across video-text tasks.

\begin{table}[!ht]
\centering
\scriptsize
\setlength{\tabcolsep}{3pt}
\begin{tabular}{llcccc}
\toprule
\textbf{Model}      & \textbf{Dataset}  & \textbf{low only} & \textbf{low+high} &\textbf{MaCP(ours)}& \textbf{4 partitions} \\
\midrule
          & MRPC              & 90.1              & 89.4     & \fs\textbf{89.7}         & 88.9 \\
  RoBERTa-B                   & CoLA              & 63.6              & 64.1     &\fs\textbf{64.6}         & 62.9 \\
                    & QNLI              & \fs\textbf{92.4}              & 92.2    & \fs\textbf{92.4}          & 92.1 \\
\midrule
             & CIFAR100          & 91.6              & \fs\textbf{91.7}          &\fs\textbf{91.7}    & 91.1 \\
                    & DTD               & 74.9              & \fs\textbf{75.9}   & \fs\textbf{75.9}           & 75.2 \\
ViT-B                      & EuroSAT           & 98.9              & 98.9   &\fs\textbf{99.1}           & 98.7 \\
                    & OxfordPets        & 93.4              & 93.3     & \fs\textbf{93.4}         & 93.1 \\
\bottomrule
\end{tabular}
\caption{Performance comparison across datasets with different feature combinations.}
\label{tab:partition}
\end{table}


\subsection{Ablation Study}

\begin{figure}
    \centering
    \includegraphics[width=\linewidth]{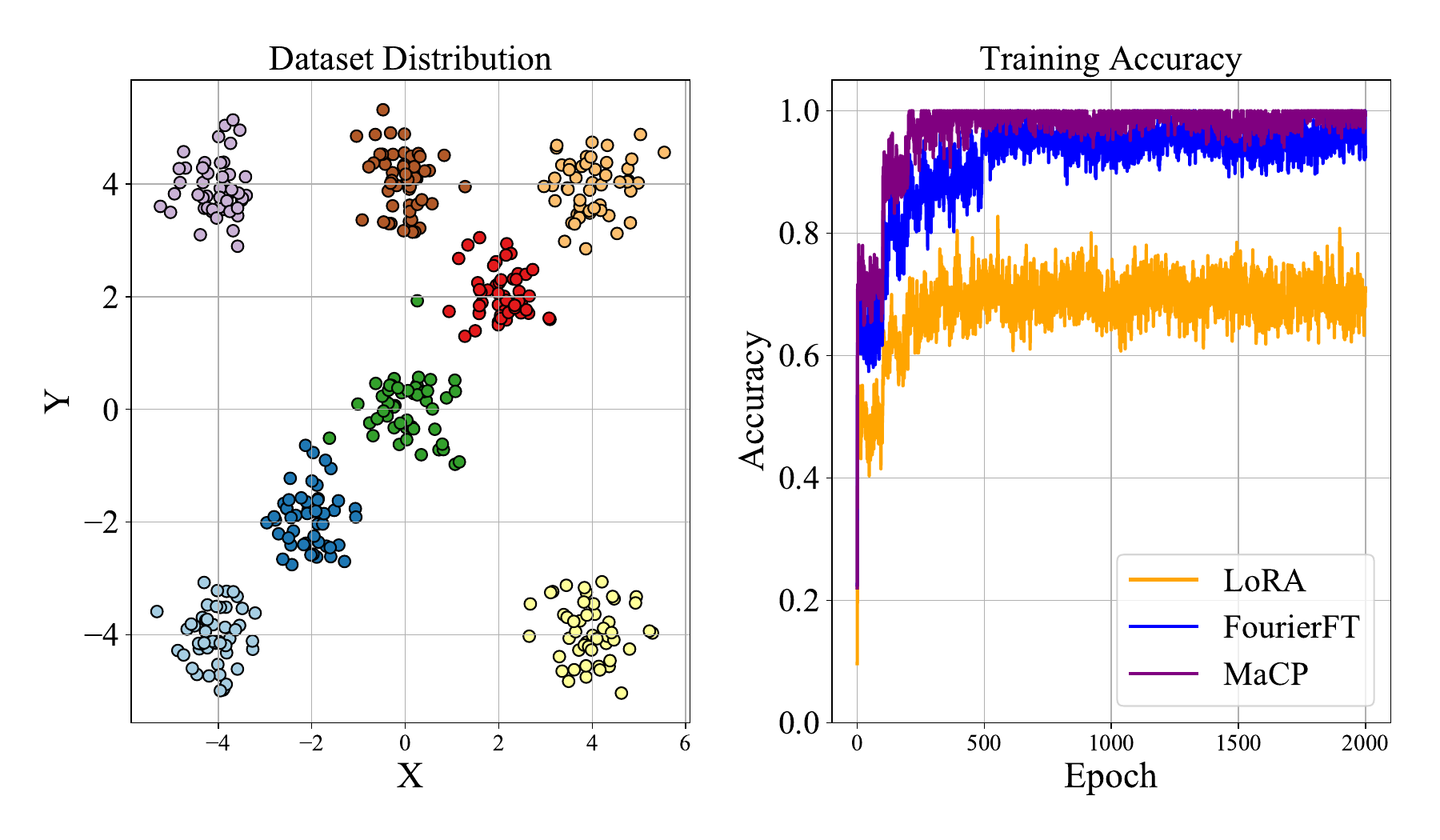}
    \caption{\small Left: Dataset distribution for the synthetic classification task, where each color represents a different class. Right: Training accuracy over 2000 epochs for LoRA, FourierFT, and MaCP. Although all methods use the same number of trainable parameters, LoRA exhibits significant oscillations and fails to converge to 100\% accuracy, indicating performance bottlenecks. FourierFT quickly reaches near-perfect accuracy within 500 epochs with stable performance. Our proposed MaCP approach outperforms both methods by converging faster (around 450 epochs) and maintaining higher accuracy with minimal fluctuations.}

    \label{fig:add}
\end{figure}

\textbf{Expressive Capability: A Comparison of MaCP, FourierFT, and LoRA.}
To intuitively evaluate the expressive power of our method, we design a simple classification task with a synthetic dataset to simulate a scenario where LoRA encounters performance bottlenecks. Specifically, we specify a 2D center point for each class of data in the 8 classes and randomly add Gaussian noise based on those points to obtain the 2D coordinates of the input. The dataset visualization is shown on the left of Figure~\ref{fig:add}. We train a single hidden layer with 64 $\times$ 64 units using LoRA ($r = 1$), FourierFT ($n = 128$), and our proposed MaCP ($n = 90$) to fit the synthesized data. Notably, all methods require the same number of trainable parameters, making it a fair comparison.

The experimental results, depicted on the right of Figure~\ref{fig:add}, highlight significant performance differences among these methods. LoRA exhibits a non-linear increase in accuracy during the initial training phase but fails to reach 100\% accuracy even after 2000 epochs. Instead, it stabilizes with considerable oscillations around 75\% accuracy, suggesting that under certain parameter constraints, LoRA encounters performance bottlenecks that hinder its ability to fully capture the underlying data distribution. In contrast, MaCP demonstrates a much smoother learning curve, reaching nearly 100\% accuracy within around 450 epochs and maintaining this performance with minimal oscillations.

\noindent \textbf{Ablation Study on Hierarchical Partitioning Schemes.}  
We conducted an ablation study with different partition schemes—“low only,” “low+high,” and “4 partitions.” Table ~\ref{tab:partition} shows that low frequencies are essential for foundational structure, but relying solely on them is insufficient, as high frequencies add crucial details, especially in texture-rich tasks like DTD. While “low+high” performs close to MaCP, ours achieves higher accuracy in tasks like EuroSAT and CoLA. The “4 partitions” scheme performs worst, suggesting that excessive partitioning dilutes key information.

\section{Conclusion}

We proposed MaCP, a hierarchical cosine projection adaptation method for fine-tuning large language models. By leveraging the energy compaction properties of the Discrete Cosine Transform (DCT), MaCP significantly reduces the number of trainable parameters while maintaining or even enhancing model performance. Extensive experiments across multiple benchmarks demonstrate that MaCP not only outperforms existing state-of-the-art methods in accuracy but also achieves superior computational efficiency and lower storage requirements. These results underscore the practical value of MaCP as a scalable and resource-efficient solution for fine-tuning large-scale models.

\section{Limitations}
This study has a few limitations. Firstly, the largest model we tested was LLaMA2-13B. Due to time constraints associated with the paper’s deadline, we could not extend our experiments to larger models, which could provide further insights into the scalability and effectiveness of MaCP. Secondly, while our method demonstrates strong performance across diverse benchmarks, we did not conduct a detailed analysis of domain-specific tasks, such as biomedical or legal text, which might require further tuning of the frequency partitioning strategy. Lastly, the computational cost of calculating the DCT, while modest, could be optimized further for extremely large-scale models.



\bibliography{acl}
\clearpage
\section*{Appendix}

In this supplementary material, we first provide detailed information about the datasets used in our experiments. Next, we outline the implementation specifics and hyper-parameter settings. We then present additional experimental results that further validate the effectiveness of the proposed MaCP method. Finally, we include examples of instruction tuning to highlight the practical application of our approach MaCP.

\section{Details of Datasets}
\paragraph{GLUE Benchmark.} 
The General Language Understanding Evaluation (GLUE) benchmark\cite{wang2019glue} is a comprehensive platform designed to evaluate and accelerate progress in natural language understanding (NLU). The benchmark comprises nine tasks that span a variety of NLU challenges such as sentiment analysis, paraphrase detection, linguistic acceptability, natural language inference, and textual similarity. Among these tasks are the Stanford Sentiment Treebank (SST-2), which focuses on binary sentiment classification for movie reviews, and the Microsoft Research Paraphrase Corpus (MRPC), which involves detecting whether two sentences are semantically equivalent. The Corpus of Linguistic Acceptability (CoLA) evaluates a model’s ability to distinguish grammatically correct sentences from incorrect ones, reflecting linguistic competence in terms of syntactic judgments. The benchmark’s diversity and inclusion of tasks with limited training data encourage the development of models that generalize well across multiple language tasks and genres.

In addition to single-sentence classification tasks, GLUE includes several sentence-pair tasks. The Question Natural Language Inference (QNLI) task is derived from the Stanford Question Answering Dataset (SQuAD). It requires models to determine if a given context sentence contains the answer to a corresponding question. The Recognizing Textual Entailment (RTE) task combines several textual entailment datasets from various domains like news and Wikipedia, testing whether a hypothesis can be logically inferred from a premise. The Semantic Textual Similarity Benchmark (STS-B) measures the similarity between sentence pairs using a regression-based approach, where models predict similarity scores on a continuous scale. 

\paragraph{E2E Benchmark.}
The E2E dataset~\cite{novikova2017e2e} is designed for training and evaluating end-to-end data-driven natural language generation (NLG) systems within the restaurant domain. It consists of over 50,000 instances and is known for its linguistic complexity, including greater lexical diversity, syntactic variation, and discourse phenomena compared to previous datasets. The evaluation is primarily conducted using five metrics: BLEU, NIST, METEOR, ROUGE-L, and CIDEr. BLEU measures the overlap of n-grams between the generated text and human references, emphasizing precision.  METEOR considers synonymy and stemming, providing a more nuanced assessment of text similarity. ROUGE-L focuses on the longest common subsequence to evaluate fluency and structure. CIDEr captures consensus by weighting n-grams based on their relevance in human references, offering a comprehensive measure of output quality.

\paragraph{Instruction Tuning Related Benchmarks}
The Alpaca dataset~\cite{taori2023stanford} is a collection of 51K instruction-following examples generated using OpenAI's text-davinci-003. It was created to fine-tune Meta's LLaMA 7B model into a lightweight, instruction-following model named Alpaca. The dataset covers a wide range of tasks, including question-answering, summarization, and classification, enabling the fine-tuned model to exhibit behavior similar to much larger models at a fraction of the cost. A specific example is as follows:

\begin{tcolorbox}[colback=green!5!white, colframe=black!75!, sharp corners, boxrule=1pt]
\footnotesize
\texttt{
\{\\
\hspace*{0.5cm} "instructions": Convert the following temperature from Celsius to Fahrenheit.\\
\hspace*{0.5cm} "input": 25 $^{\circ}$C\\
\hspace*{0.5cm} "output": 25 $^{\circ}$C is equal to 77$^{\circ}$F.\\
\}
}
\end{tcolorbox}
MT-Bench~\cite{zheng2024judging} is a recently introduced benchmark designed to evaluate the instruction-following capabilities of language foundation models. It consists of a series of open-ended questions aimed at assessing model performance across diverse aspects such as writing, roleplay, reasoning, mathematics, coding, information extraction, STEM, and the humanities. MT-Bench effectively distinguishes these abilities through tailored questions, providing a more comprehensive evaluation. A specific example from the benchmark is provided below.

\begin{tcolorbox}[colback=green!5!white, colframe=black!75!, sharp corners, boxrule=1pt]
\footnotesize
\texttt{
\{\\
\hspace*{0.5cm} "Q1": What is the square root of 144?\\
\hspace*{0.5cm} "Q2(follow-up)": If you multiply the answer by 2, what is the result?\\
\hspace*{0.5cm} "Solution": Q1: The square root is 12. Q2: The result is 24.\\
\}
}
\end{tcolorbox}

Vicuna Eval~\cite{chiang2023vicuna} is a benchmark designed to assess the alignment of large language models (LLMs) with human preferences and serves as the predecessor to MT-Bench. Vicuna Eval evaluates models across a diverse set of topics, including coding, writing, mathematics, counterfactual reasoning, Fermi estimation, common sense, roleplay, knowledge, and generic tasks. It provides a comprehensive framework for understanding how well models align with human expectations in varied scenarios. A specific example from this evaluation is presented below.

\begin{tcolorbox}[colback=green!5!white, colframe=black!75!, sharp corners, boxrule=1pt]
\small
\texttt{
\{\\
\hspace*{0.5cm} "question": Explain the concept of recursion with a simple example.\\
\hspace*{0.5cm} "category": programming.\\
\}
}
\end{tcolorbox}

\paragraph{Image Classification Datasets}
Tab.~\ref{tab-image} provides comprehensive details about four widely-used vision datasets: CIFAR100, DTD, EuroSAT, and OxfordPets. It outlines key information such as the number of training (\#Train), validation (\#Val), and test (\#Test) samples, along with the number of classes (\#Class) in each dataset. These datasets span diverse domains, from object recognition (CIFAR100~\cite{krizhevsky2009learning}) to texture classification (DTD~\cite{cimpoi2014describing}), satellite image classification (EuroSAT~\cite{helber2019eurosat}), and pet identification (OxfordPets~\cite{parkhi2012cats}). The variety in datasets ensures that the models are evaluated across different visual tasks, offering a robust assessment of their performance.

For consistency in model evaluation, all datasets are rescaled to a resolution of 224 × 224. This standardized input size simplifies comparisons by ensuring that all models receive uniformly sized images, crucial for fair benchmarking. The datasets vary in size and complexity, with CIFAR100 having the largest number of samples (60,000) across 100 classes, while OxfordPets focuses on a more specialized task with 37 classes. The information in this table highlights the diversity of the datasets, each contributing unique challenges to model evaluation.

\begin{table}
\centering
\resizebox{0.47\textwidth}{!}{%
\begin{tabular}{l|r|r|r|r|l}
\toprule
\textbf{Dataset} & \textbf{\#Train} & \textbf{\#Val} & \textbf{\#Test} & \textbf{\#Class} & \textbf{Rescaled res.} \\
\midrule
CIFAR100  & 45,000 & 5,000 & 10,000 & 100 & \multirow{4}{*}{224 $\times$ 224} \\
DTD  & 4,060 & 452 & 1,128 & 47 &  \\
EuroSAT & 16,200 & 5,400 & 5,400 & 10 &  \\
OxfordPets & 3,312 & 368 & 3,669 & 37 & \\
\bottomrule
\end{tabular}%
}
\caption{Details about the vision datasets.}
\label{tab-image}
\end{table}

\section{Hyperparamaters}

\begin{table*}
\centering
\resizebox{0.7\textwidth}{!}{%
\begin{tabular}{ll|ccccccc}
\toprule
\textbf{Model} & \textbf{Hyperparameter} & \textbf{STS-B} & \textbf{RTE} & \textbf{MRPC} & \textbf{CoLA} & \textbf{SST-2} & \textbf{QNLI} \\
\midrule
\multirow{7}{*}{\textbf{Both}} 
& Optimizer & \multicolumn{6}{c}{AdamW} \\
& LR Schedule & \multicolumn{6}{c}{Linear} \\
& Warmup Ratio & \multicolumn{6}{c}{0.06} \\
& Frequency Bias & \multicolumn{6}{c}{False} \\
& $n_{MaCP}$ & \multicolumn{6}{c}{700} \\
& $n_{FourierFT}$ & \multicolumn{6}{c}{1000} \\
& Seeds & \multicolumn{6}{c}{\{0, 11111, 22222, 33333, 44444\}} \\
\midrule
\multirow{7}{*}{\textbf{Base}} 
& Epochs & 60 & 90 & 30 & 100 & 40 & 40 \\
& Learning Rate (MaCP) & 9E-2 & 9E-2 & 5E-2 & 1.2E-1 & 5E-2 & 1E-2 \\
& Learning Rate (FourierFT) & 9E-2 & 9E-2 & 5E-2 & 1.2E-1 & 5E-2 & 1E-2 \\
& Learning Rate (VeRA) & 9E-2 & 9E-2 & 5E-2 & 1.2E-1 & 5E-2 & 1E-2 \\
& Learning Rate (Head) & 9E-3 & 1.1E-2 & 6E-3 & 8E-3 & 6E-3 & 1E-3 \\
& Max Seq. Len & 512 & 512 & 512 & 512 & 512 & 512 \\
& Scaling Value & 84 & 110 & 141 & 49 & 140 & 29 \\
& Batch Size & 32 & 32 & 32 & 32 & 32 & 32 \\
\midrule
\multirow{7}{*}{\textbf{Large}} 
& Epochs & 30 & 60 & 30 & 80 & 10 & 30 \\
& Learning Rate (MaCP) & 7E-2 & 8E-2 & 6E-2 & 4.3E-2 & 4.3E-2 & 6E-2 \\
& Learning Rate (FourierFT) & 7E-2 & 8E-2 & 6E-2 & 4.3E-2 & 4.3E-2 & 6E-2 \\
& Learning Rate (VeRA) & 7E-2 & 8E-2 & 6E-2 & 4.3E-2 & 4.3E-2 & 6E-2 \\
& Learning Rate (Head) & 1E-3 & 5E-3 & 1E-3 & 1.1E-2 & 1E-3 & 5E-3 \\
& Max Seq. Len & 512 & 512 & 512 & 256 & 128 & 512 \\
& Scaling Value & 121 & 90 & 120 & 90 & 69 & 69 \\
& Batch Size & 32 & 32 & 32 & 128 & 32 & 32 \\
\bottomrule
\end{tabular}%
}
\caption{\small Hyperparameters used for MaCP across various GLUE tasks.}
\label{tab:nluh}
\end{table*}

\paragraph{Hyperparameters on GLUE benchmarks}Tab.~\ref{tab:nluh} describes the key hyperparameters used in the experiments across different GLUE tasks and model sizes (Base and Large). The table outlines the learning rate schedules, optimizer settings, warmup ratios, and seed values for reproducibility. For both Base and Large models, the AdamW optimizer is used with a linear learning rate schedule and a warmup ratio of 0.06. The frequency bias is set to false, and the frequency coefficient \( n \) is fixed at 700 for MaCP unless specified otherwise. Each experiment is run with 5 different seeds \{0, 11111, 22222, 33333, 44444\}.

For the Base models, the number of training epochs ranges from 30 to 100 depending on the task, with SST-2 requiring the longest training time. The FourierFT and MaCP methods use a higher learning rate for the base models compared to the learning rate used for fine-tuning the head layers. 

In contrast, the Large models typically require fewer epochs but use slightly lower learning rates. The batch size remains consistent across both model sizes, set at 32 for all tasks. Additionally, max sequence lengths are adapted to fit the needs of each task, with longer sequences allocated for more complex tasks like CoLA and QNLI.

\begin{table}
\centering
\resizebox{0.4\textwidth}{!}{%
\begin{tabular}{l|cc}
\toprule
\textbf{Hyperparameter} & \textbf{Medium} & \textbf{Large} \\
\midrule
Optimizer & \multicolumn{2}{c}{AdamW} \\
Learning Rate (MaCP) & 2E-2 & 5E-2 \\
Learning Rate (FourierFT) & 2E-2 & 5E-2 \\
Learning Rate (VeRA) & 2E-2 & 5E-2 \\
Learning Rate (Head) & 2E-4 & 1E-4 \\
Batch Size & \multicolumn{2}{c}{128} \\
Weight Decay & 0.01 & 0.03 \\
$n_{MaCP}$ &\multicolumn{2}{c}{700} \\
$n_{FourierFT}$ & \multicolumn{2}{c}{1000} \\
Scaling value $\alpha$ & \multicolumn{2}{c}{300} \\
Epochs & \multicolumn{2}{c}{5}  \\
Label Smooth & \multicolumn{2}{c}{0.1}  \\
LR Schedule & \multicolumn{2}{c}{Linear} \\
\bottomrule
\end{tabular}%
}
\caption{\small Hyperparameter settings on E2E benchmark}
\label{tab:nlgh}
\end{table}

\paragraph{Hyperparameter settings on E2E benchmark} Tab.~\ref{tab:nlgh} outlines the hyperparameter configurations used for the medium and large models on the E2E benchmark. Both models are optimized using AdamW with a linear learning rate schedule. The learning rates for MaCP and FourierFT are set to \(2E-2\) for the medium model and \(5E-2\) for the large model, while the head layers have lower learning rates of \(2E-4\) and \(1E-4\), respectively. Both models use a batch size of 128, with weight decay values of 0.01 and 0.03 for the medium and large models, respectively. The number of selected frequencies, \(n\), is set to 700 for MaCP and 1000 for FourierFT, while the scaling value \(\alpha\) is fixed at 300 for both models. Additionally, label smoothing is applied with a value of 0.1, and training is conducted for 5 epochs.

\begin{table}
\centering
\resizebox{0.45\textwidth}{!}{%
\begin{tabular}{l|cccc}
\toprule
\textbf{Hyperparameter} & \textbf{LoRA} & \textbf{FourierFT} & \textbf{MaCP} & \textbf{VeRA}\\
\midrule
Optimizer & \multicolumn{4}{c}{AdamW} \\
Warmup Ratio & \multicolumn{4}{c}{0.06} \\
Batch Size & \multicolumn{4}{c}{4} \\
Acc. Steps & \multicolumn{4}{c}{4} \\
Epochs & \multicolumn{4}{c}{1 or 2} \\
$n$ &  -- & 1000 & 700 & --\\
Scaling Value $\alpha$ & 300.0 & 16.0 & 16.0 & 300.0\\
LR Schedule & \multicolumn{4}{c}{Linear} \\
Learning Rate & 3E-2 & 3E-3 & 3E-3 &  3E-3 \\
\bottomrule
\end{tabular}%
}
\caption{\small Hyperparameter settings for instruction-tuning  configurations.}
\label{tab:hyperparamsIn}
\end{table}

\paragraph{Hyperparameter Setup for Instruction-Tuning} Table~\ref{tab:hyperparamsIn} summarizes the key hyperparameters used for fine-tuning the LoRA, FourierFT, and MaCP models. Across all methods, the optimizer is AdamW, with a warmup ratio of 0.06, a batch size of 4, and gradient accumulation steps set to 4, providing stability during training. The default training duration is 1 epoch, with certain experiments—such as the motivation example in the introduction and the ablation study in the supplementary material—using 2 epochs where indicated.

For MaCP, the parameter \( n \) is set to 700. The scaling value \( \alpha \) differs across methods: it is 300.0 for LoRA, 16.0 for FourierFT, and also 16.0 for MaCP. The learning rate is tuned individually, with LoRA using 3E-2 and both FourierFT and MaCP using a lower rate of 3E-3. All methods follow a linear learning rate schedule.

\paragraph{Hyperparameter setup for image classification}

\begin{table}
\centering
\resizebox{0.47\textwidth}{!}{%
\begin{tabular}{l|cccc}
\toprule
\textbf{Hyperparameter} & \textbf{CIFAR100} & \textbf{DTD} & \textbf{EuroSAT} &  \textbf{OxfordPets} \\
\midrule
Epochs & \multicolumn{4}{c}{10} \\
Optimizer & \multicolumn{4}{c}{AdamW} \\
LR Schedule & \multicolumn{4}{c}{Linear} \\
$n_{MaCP}$ & \multicolumn{4}{c}{2400} \\
$n_{FourierFT}$ & \multicolumn{4}{c}{3000} \\
$\alpha$ & \multicolumn{4}{c}{300.0} \\
Learning Rate (MaCP) & 2E-1 & 3E-1 & 2E-1 & 3E-1  \\
Learning Rate (FourierFT) & 2E-1 & 3E-1 & 2E-1 & 3E-1  \\
Learning Rate (VeRA) & 2E-1 & 3E-1 & 2E-1 & 3E-1  \\
Learning Rate (Head) & 7E-4 & 1E-3 & 8E-4 & 1E-3  \\
Weight Decay & 1E-4 & 7E-5 & 3E-4 & 8E-4 \\
\bottomrule
\end{tabular}%
}
\caption{\small Hyperparameter setup for image classification.}
\label{tab:MaCP_image}
\end{table}

Tab.~\ref{tab:MaCP_image} presents the hyperparameter configurations used for fine-tuning on the CIFAR100, DTD, EuroSAT, and OxfordPets datasets for image classification tasks. The table lists the common settings across these datasets, including the use of the AdamW optimizer, a linear learning rate schedule, and a consistent training setup of 10 epochs. The number of frequency components ($n$) is set to 2400 for MaCP and 3000 for FourierFT across all datasets.

For MaCP and FourierFT, the learning rate varies slightly across datasets, ranging from 2E-1 to 3E-1, while the learning rate for the classification head ranges between 7E-4 and 1E-3. The weight decay is tuned for each dataset, with values ranging from 7E-5 to 1E-4 for DTD and CIFAR100, and slightly higher at 3E-4 and 8E-4 for EuroSAT and OxfordPets, respectively.

\paragraph{Hyperparameter setup for video-text understanding}
Table~\ref{tab:SSH_video} summarizes the hyperparameter configurations used for fine-tuning the models across four video-text understanding tasks: TVQA, How2QA, TVC, and YC2C. The training process was consistent across all tasks, with each model trained for 7 epochs using the AdamW optimizer. A common learning rate schedule (\texttt{Q,K}) was employed for all models.

For the frequency-based methods, the number of selected frequency components, $n$, was set to 4500 for MaCP and 6000 for FourierFT. For the LoRA and DoRA methods, the rank $r$ was set to 128 for both. The scaling factor $\alpha$ was fixed at 300.0 across all methods. 

The learning rate for MaCP, FourierFT, and VeRA was kept at 3E-4 for all tasks, with slight adjustments in the weight decay for different datasets: 2E-3 for TVQA, TVC, and YC2C, and 2E-5 for How2QA.

\begin{table}
\centering
\resizebox{0.47\textwidth}{!}{%
\begin{tabular}{l|cccc}
\toprule
\textbf{Hyperparameter} & \textbf{TVQA} & \textbf{How2QA} & \textbf{TVC} &  \textbf{YC2C} \\
\midrule
Epochs & \multicolumn{4}{c}{7} \\
Optimizer & \multicolumn{4}{c}{AdamW} \\
LR Schedule & \multicolumn{4}{c}{Q,K} \\
$n_{MaCP}$ & \multicolumn{4}{c}{4500} \\
$n_{FourierFT}$ & \multicolumn{4}{c}{6000} \\
$r_{LoRA}$ & \multicolumn{4}{c}{128} \\
$r_{DoRA}$ & \multicolumn{4}{c}{128} \\
$\alpha$ & \multicolumn{4}{c}{300.0} \\
Learning Rate (MaCP) & 3E-4 & 3E-4 & 3E-4 & 3E-4  \\
Learning Rate (FourierFT) & 3E-4 & 3E-4 & 3E-4 & 3E-41  \\
Learning Rate (VeRA) & 3E-4 & 3E-4 & 3E-4 & 3E-4 \\
Weight Decay & 2E-3 & 2E-5 & 2E-3 & 2E-3 \\
\bottomrule
\end{tabular}%
}
\caption{\small Hyperparameter setup for video-text understanding}
\label{tab:SSH_video}
\end{table}

\subsection{Additional Ablation Study}
This ablation study further examines the performance of MaCP when trained for two epochs, with results shown in Table~\ref{tab:mtbench_vicuna_ab}. Even with extended training, our method consistently surpasses other approaches in both efficiency and performance. For instance, in the LLaMA2-7B model, MaCP achieves an MT-Bench score of 5.27 and a Vicuna score of 7.51 using only 0.045M parameters. In comparison, FourierFT, which uses more parameters (0.064M), falls short in performance. Similarly, on the LLaMA2-13B model, MaCP outperforms all other methods, recording the highest scores of 5.97 on MT-Bench and 8.01 on Vicuna with only 0.056M trainable parameters, compared to FourierFT's 0.08M. These results, highlighted in Table~\ref{tab:mtbench_vicuna_ab}, clearly indicate that MaCP remains robust and effective, solidifying its advantage in limited computational resources.

\begin{table}
\centering
\resizebox{0.43\textwidth}{!}{%
\begin{tabular}{l|l|c|crcc}
\toprule
\textbf{Model} & \textbf{Method} & \textbf{\# Tr. Para.} & \textbf{MT-Bench} & \textbf{Vicuna} \\
\midrule
\multirow{5}{*}{\textbf{LLaMA2-7B}} 
& LoRA & 159.9M & 5.24 & 7.38 \\
& DoRA & 163.7M & 5.25 & 7.41 \\
& VeRA & 1.6M & 5.23 & 7.47 \\
& FourierFT & 0.064M & 5.19 & 7.50 \\
& \textbf{MaCP} & \fs\textbf{0.045M} & \fs\textbf{5.27} & \fs\textbf{7.51}\\
\midrule
\multirow{5}{*}{\textbf{LLaMA2-13B}} 
& LoRA & 250.3M & 5.85 & 7.89\\
& DoRA & 264.5M & 5.85 & 7.91 \\
& VeRA & 2.4M & 5.96 & 7.86 \\
& FourierFT & 0.08M & 5.89 & 7.92 \\
& \textbf{MaCP} & \fs\textbf{0.056M} & \fs\textbf{5.97} & \fs\textbf{8.01} \\
\midrule
\multirow{5}{*}{\textbf{LLaMA3.1-8B}} 
& LoRA & 183.3M & 5.69 & 7.52 \\
& DoRA & 186.9M & 5.72 & 7.59 \\
& VeRA & 1.9M & 5.67 & 7.49 \\
& FourierFT & 0.073M & 5.70 & 7.67 \\
& \textbf{MaCP} & \fs\textbf{0.05M} & \fs\textbf{5.73} & \fs\textbf{7.72} \\
\bottomrule
\end{tabular}%
}
\caption{\small Performance comparison of various fine-tuning methods on LLaMA models (LLaMA2-7B, LLaMA2-13B, and LLaMA3.1-8B) evaluated using MT-Bench and Vicuna scores. All models are fine-tuned for 2 epochs and assessed by GPT-4. The scores represent the average across multiple trials, where higher scores indicate better performance. Notably, MaCP consistently outperforms other methods with significantly fewer trainable parameters.}
\label{tab:mtbench_vicuna_ab}
\end{table}

\section{Examples of Instruction Tuning}
This section highlights practical examples generated by the LLaMA3.1-8B model fine-tuned with FourierFT and MaCP, illustrating the comparative effectiveness of each method. FourierFT uses 146K trainable parameters, while MaCP requires only 100K. Each case includes a prompt, the answers generated by both models, and a review of the responses assessed by GPT-4.

This section presents three case studies comparing models fine-tuned with FourierFT and MaCP. In a restaurant review, MaCP provided more detailed and engaging insights, while FourierFT was concise but less descriptive (Tab.~\ref{tab:res}). In coding instructions for finding common elements in arrays, both models were correct, but MaCP’s explanation was more comprehensive and user-friendly (Tab.~\ref{tab:code}). Lastly, in film analysis, MaCP offered a deeper, more nuanced critique of plot, character development, and directorial choices, outperforming FourierFT’s straightforward evaluation (Tab.~\ref{tab:film}).

\begin{table*}[ht]
\centering
\resizebox{0.9\textwidth}{!}{%
\begin{tabular}{c|p{7cm}|p{7cm}}
\toprule
\textbf{Method} & \textbf{FourierFT} & \textbf{MaCP} \\
\midrule
\textbf{Prompt} & \multicolumn{2}{c}{Write a restaurant review, focusing on the food quality, service, and overall ambiance.} \\
\midrule
\textbf{Answer} & 
The restaurant offered a delightful dining experience with an excellent variety of dishes. The flavors were well-balanced, and the ingredients fresh, though some of the main courses lacked the expected depth. The service was prompt, but it felt somewhat impersonal at times. The ambiance was warm and inviting, with tasteful decor that complemented the overall experience. Overall, it was a pleasant visit with a few areas for improvement. &
The restaurant experience was nothing short of exceptional. From the moment we walked in, the attentive staff made sure we felt welcomed. The food was outstanding, with each dish crafted with precision and bursting with flavor. The standout was the chef’s special, which delivered a perfect combination of taste and presentation. The ambiance was charming, offering a cozy yet elegant atmosphere. Every detail was carefully thought out, making it a memorable evening overall. \\
\midrule
\textbf{Review} & 
Assistant 1 captured the essence of the dining experience but could have elaborated more on specific dishes or details about the restaurant’s atmosphere. The review was concise and clear but left out some critical aspects that would make it more informative and engaging. &
Assistant 2 provided a thorough review, highlighting not just the food quality but also the service and ambiance. The attention to detail, such as mentioning the standout dish and how the atmosphere enhanced the overall experience, added depth and made the review more compelling and enjoyable to read. \\
\midrule
\textbf{Score} & 7.65 & \textbf{7.80} \\
\bottomrule
\end{tabular}%
}
\caption{Comparison of generated responses from models fine-tuned with FourierFT and MaCP on a restaurant review prompt.}
\label{tab:res}
\end{table*}

\begin{table*}[ht]
\centering
\resizebox{0.95\textwidth}{!}{%
\begin{tabular}{p{2cm}|p{7.5cm}|p{7.5cm}}
\toprule
\textbf{Method} & \textbf{FourierFT} & \textbf{MaCP} \\
\midrule
\textbf{Prompt} & \multicolumn{2}{p{15cm}}{Write a Python program to find the common elements in two arrays without using extra data structures. Provide a step-by-step explanation.} \\
\midrule
\textbf{Answer} & 
\textbf{Step-by-Step Explanation:} The process starts by sorting both arrays to simplify the comparison. Sorting helps identify common elements more efficiently. Next, two pointers are initialized, one for each array, starting at the beginning. The algorithm then traverses both arrays in parallel. When the elements at both pointers are equal, the common element is added to the result list. If the first array’s element is smaller, the pointer in that array is moved forward. Otherwise, the pointer in the second array is advanced. This continues until one of the pointers exceeds the array’s length, at which point all common elements have been found. \textbf{Pseudo-code:} Sort(arr1), Sort(arr2). Initialize i, j = 0. While i < len(arr1) and j < len(arr2): If arr1[i] == arr2[j]: Add to result. Else if arr1[i] < arr2[j]: i++. Else: j++. Return result. & 
\textbf{Step-by-Step Explanation:} Begin by sorting both arrays, allowing for easier comparison during traversal. Then, use two pointers to scan both arrays from the start. As the traversal progresses, if the current elements match, they are added to the output list. If the element in the first array is smaller, its pointer is moved forward. If the element in the second array is smaller, its pointer advances. The algorithm concludes when either pointer reaches the end, signaling that all possible common elements have been identified. \textbf{Pseudo-code:} Sort(arr1), Sort(arr2). Initialize i, j = 0. Loop until either pointer is out of range: If arr1[i] == arr2[j]: Add to result. If arr1[i] < arr2[j]: Move i forward. Else: Move j forward. Output the result. \\
\midrule
\textbf{Review} & 
This explanation is clear and intuitive, making the approach easy to follow. The importance of sorting and using pointers is well-highlighted. However, while the explanation is technically sound, it could benefit from addressing potential edge cases like handling empty arrays or duplicates. & 
This explanation is detailed and user-friendly, with a strong emphasis on clear decision-making within the algorithm. The pointers are well-explained, and the pseudo-code is both concise and logical. The approach effectively handles typical scenarios, but more depth on edge cases could further improve it. \\
\midrule
\textbf{Score} & 7.79 & \textbf{7.97} \\
\bottomrule
\end{tabular}
}
\caption{Comparison of generated responses from models fine-tuned with FourierFT and MaCP on a coding instruction prompt.}
\label{tab:code}
\end{table*}

\begin{table*}[ht]
\centering
\resizebox{0.9\textwidth}{!}{%
\begin{tabular}{c|p{7.7cm}|p{7.7cm}}
\toprule
\textbf{Method} & \textbf{FourierFT} & \textbf{MaCP} \\
\midrule
\textbf{Prompt} & \multicolumn{2}{p{15cm}}{Provide a detailed analysis of the film’s plot structure, character development, directorial choices, and the role of cinematography in reinforcing the central themes.} \\
\midrule
\textbf{Answer} & 
The film expertly weaves together a multi-layered plot that explores themes of redemption, identity, and the consequences of past actions. The narrative unfolds in a non-linear fashion, employing flashbacks and parallel timelines that gradually reveal the backstory of the characters. This intricate structure keeps the audience engaged, constantly piecing together clues while deepening the emotional stakes. The character development is a highlight, particularly the protagonist, whose transformation from a morally ambiguous figure to someone grappling with guilt and seeking redemption is portrayed with remarkable depth. The supporting characters, though less developed, provide crucial counterpoints that enhance the protagonist’s journey. The director’s use of long takes and carefully framed shots adds a sense of intimacy and tension, making the audience feel like silent observers within the story. The cinematography further complements this by using stark contrasts between light and shadow to symbolize the internal conflicts of the characters. The film’s visual language is rich with symbolism, from recurring motifs like broken mirrors to the deliberate use of desaturated color palettes during scenes of despair. However, the third act, while visually stunning, slightly overextends itself, relying too heavily on exposition. Despite this, the film’s ability to blend a complex narrative with powerful imagery makes it a standout example of modern cinema. &
The film’s narrative is a compelling exploration of complex human emotions, weaving a story that is both introspective and expansive in scope. The plot structure is intricately layered, using a blend of real-time events and flashbacks to gradually reveal the true motivations behind each character’s actions. The protagonist’s arc is especially well-crafted, showing a slow but steady transformation driven by a combination of external conflicts and internal struggles. This gradual change is mirrored in the pacing, which starts off deliberately slow, allowing the audience to fully grasp the gravity of each decision made by the characters. The secondary characters, while often playing subtle roles, are given moments that add depth to the overall narrative. Directorial choices play a significant role in maintaining the film’s atmospheric tone. Through the use of visual storytelling, the director skillfully conveys the tension and unresolved emotions between characters without resorting to overt dialogue. The cinematography is not just a backdrop but an active participant in the storytelling. For example, the frequent use of close-up shots during moments of silence emphasizes the unspoken emotions that linger beneath the surface. The decision to film key scenes during golden hour creates a visual contrast between the warmth of the setting sun and the cold, stark reality of the characters’ situations. The careful choreography of camera movements during action sequences adds a layer of realism that pulls the audience further into the narrative. While the film’s conclusion leaves some plot threads unresolved, this ambiguity feels intentional, prompting the viewer to reflect on the underlying themes long after the credits roll. \\
\midrule
\textbf{Review} & 
The analysis provided a thorough overview of the film’s plot structure and character development. While it accurately identified key themes and offered insight into directorial choices, it occasionally lacked specific examples that would make the points more concrete. The discussion on cinematography was solid, but a more detailed breakdown of particular scenes or camera techniques could have strengthened the overall argument. Overall, this review was clear, relevant, and informative, but some areas could benefit from more in-depth analysis and illustrative examples. &
This analysis delivered a comprehensive breakdown of the film’s narrative layers, seamlessly integrating observations on character development, pacing, and the director’s stylistic choices. The reviewer highlighted the interaction between visual storytelling and thematic content, using specific examples to support each point. The detailed exploration of the protagonist’s transformation was particularly effective, demonstrating a nuanced understanding of the film’s emotional core. The cinematography discussion not only described the techniques used but contextualized their impact within the broader narrative framework. The thoroughness and attention to detail in this review made it both insightful and engaging. \\
\midrule
\textbf{Score} & 7.91 & \textbf{8.09} \\
\bottomrule
\end{tabular}%
}
\caption{Comparison of generated responses from models fine-tuned with FourierFT and MaCP on an extended movie analysis prompt.}
\label{tab:film}
\end{table*}

\end{document}